\documentclass{article}
\usepackage{ijcai25}

\usepackage{times}
\usepackage{soul}
\usepackage{url}
\usepackage[hidelinks]{hyperref}
\usepackage[utf8]{inputenc}
\usepackage[small]{caption}
\usepackage{graphicx}
\usepackage{amsmath}
\usepackage{amsthm}
\usepackage{booktabs}
\usepackage{algorithm}
\usepackage{algorithmic}
\usepackage[switch]{lineno}
\usepackage{amsfonts}
\usepackage{makecell}
\usepackage{siunitx}
\usepackage{listings}
\usepackage{wrapfig}
\usepackage{tabularx} 
\usepackage{textcomp}
\usepackage{tikz}
\usepackage{array}
\usepackage{subfig}
\usepackage{booktabs}
\usepackage{float}
\usepackage{multirow}
\usepackage{bm}

\title{Beyond Local Selection: Global Cut Selection for\\ Enhanced Mixed-Integer Programming}

\author{
Shuli Zeng
\and
Sijia Zhang\and
Shaoang Li\and
Feng Wu\And
Xiang-Yang Li$^*$\\
\affiliations
School of Computer Science and Technology, University of Science and Technology of China\\
\emails
\{zengshuli0130,sxzsj,lishaoa\}@mail.ustc.edu.cn,
\{wufeng02,xiangyangli\}@ustc.edu.cn
}

\begin{document}

\maketitle

\begin{abstract}
In mixed-integer programming (MIP) solvers, cutting planes are essential for Branch-and-Cut (B\&C) algorithms as they reduce the search space and accelerate the solving process. Traditional methods rely on hard-coded heuristics for cut plane selection but fail to leverage problem-specific structural features. Recent machine learning approaches use neural networks for cut selection but focus narrowly on the efficiency of single-node within the B\&C algorithm, without considering the broader contextual information. To address this, we propose Global Cut Selection (GCS), which uses a bipartite graph to represent the search tree and combines graph neural networks with reinforcement learning to develop cut selection strategies. Unlike prior methods, GCS applies cutting planes across all nodes, incorporating richer contextual information. Experiments show GCS significantly improves solving efficiency for synthetic and large-scale real-world MIPs compared to traditional and learning-based methods.
\end{abstract}

\section{Introduction}

Mixed-integer programming (MIP) is a widely used optimization framework for solving various important real-world applications, such as supply chain management~\cite{paschos2014applications}, production planning~\cite{junger200950} and scheduling~\cite{chen2010integrated}. Solving MIPs is very challenging because they are NP-hard problems~\cite{bixby1992implementing}. 
State-of-the-art MIP solvers, such as Gurobi~\cite{gurobi} and SCIP~\cite{bestuzheva2021scip}, typically employ branch-and-bound algorithms~\cite{land2010automatic}. A key technique for improving the efficiency of B\&B is the use of cutting planes, which help prune the search space by tightening LP relaxations. 
Research on cutting planes can be categorized into two primary areas: cut generation~\cite{guyon2010cut} and cut selection~\cite{turner2022adaptive}. Cut generation involves creating valid linear inequalities that tighten the LP relaxations~\cite{achterberg2007constraint}. However, adding all generated cuts incurs substantial computational costs~\cite{wesselmann2012implementing}. To address this, cut selection strategies have been developed to choose a suitable subset of the generated cuts~\cite{wesselmann2012implementing}. Our paper focuses on the cut selection problem, which plays a crucial role in the overall performance of the solver~\cite{achterberg2007constraint,tang2020reinforcement,paulus2022learning}.

MIP solvers commonly rely on expert-designed, hard-coded heuristics~\cite{turner2023cutting} for selecting cutting planes. However, such heuristics often fail to fully exploit the specific structural features of the given integer programming problems, such as production scheduling and packing. To further enhance the efficiency of MIP solvers, recent methods propose to use machine learning to learn cut selection strategies~\cite{wang2023learning}. These methods offer promising approaches to learning more effective methods by capturing latent features from specific types of problems. 

Previous learning-based research~\cite{tang2020reinforcement,huang2022learning,wang2023learning} in this area has primarily concentrated on local strategies, focusing on the selection of cutting planes at individual node in search tree. However, global decisions, such as determining which nodes should receive which cutting planes and when to apply them throughout the entire search process, have largely been overlooked. The study by Berthold et al.~\shortcite{berthold2022learning} shows that choosing between adding cutting planes only at the root node or at all nodes can improve the SCIP solver's performance by approximately 11\% on the MIPLIB 2017 dataset~\cite{gleixner2021miplib}. Because local strategies, such as score-based policies~\cite{tang2020reinforcement,huang2022learning} or hierarchical network selections~\cite{wang2023learning}, fail to leverage more global information, they can limit the overall efficiency of the selection strategy, particularly for large-scale problems.


Through our experiments, we observed that simply extending a learning-based cut selection strategy from the root node to all nodes can lead to a 30\% decrease in performance. Furthermore, as the problem size grows, the performance gap becomes increasingly pronounced.  We posit that effective cut selection requires a strategy that considers the entire search tree. Learning at the individual node level often fails to capture inter-node dependencies and global search tree structure, both of which are critical for optimal decision-making. Moreover, without a global representation, it becomes difficult to track the history of previously selected cutting planes. This limitation is significant because the effects of cutting planes are not independent; their combined impact on the search process can vary substantially. Thus, a lack of historical and contextual information hinders the ability to identify combinations of cuts that work well together. 

In principle, a branch-and-cut algorithm could leverage information from the entire search tree for cut selection. However, a search tree can easily consist of thousands of nodes, generating an immense and often redundant volume of data that is challenging to represent in a single model. Consequently, prior works~\cite{achterberg2007constraint,baltean2019scoring,huang2022learning,paulus2022learning} have typically limited their focus to information from the current node alone. For example, Wang et al.~\shortcite{wang2023learning} manually extracted 13-dimensional features based on one cut and the root node. This approach, however, is insufficient for effective cut selection across all nodes, as it lacks a broader context of the search tree. The resulting strategy treats each node in isolation, akin to the root node, which can hinder performance.

To address the challenges in cut selection, we propose a new approach to represent global tree information tailored to the problem of cutting plane selection. Specifically, we introduce the Global Cut Selection (GCS) framework, which incorporates innovative strategies for feature extraction and decision-making.  First, the method represents the entire search tree using all leaf nodes. By analyzing the common and distinct parts of each leaf node, it embeds the information from all leaf nodes into a single graph. Additionally, GCS tracks the history of each added cutting plane on the graph, allowing the network to make more informed decisions based on past data. A Graph Neural Network (GNN) then aggregates this information to select cutting planes, while a Transformer network considers the relationships between all cutting planes for optimal selection. This approach ensures that the selection process leverages the full context of the search tree, leading to more effective and efficient cutting plane decisions.


The contributions of this paper are summarized as follows:

 \textbullet~ We propose GCS, a reinforcement learning (RL) framework for adding cutting planes at all nodes, and empirically demonstrate its advantages over single-node selection methods.

 \textbullet~ GCS leverages the information of the entire search tree to inform cutting plane selection. GCS represents the search tree as a bipartite graph, which serves as the input to the network, rather than relying solely on the information from the current node as in previous work.

\textbullet~ We conducted extensive experiments on both small and large datasets, demonstrating that GCS consistently outperforms previous state-of-the-art (SOTA) strategies in terms of solving speed, node reduction, and generality.
\section{Background}
\textbf{Mixed integer programming (MIP).}
The general form of a MIP is given by
\begin{equation}
z^* = \min_{\mathbf{x}} \left\{\mathbf{c}^T \mathbf{x} \mid \mathbf{A} \mathbf{x} \leq \mathbf{b}, \mathbf{x} \in \mathbb{R}^n, x_j \in \mathbb{Z}, \forall j \in I \right\}
\label{def_milp}
\end{equation}
where $\mathbf{x} \in \mathbb{R}^{n}$ is the set of $n$ decision variables, $\mathbf{c} \in \mathbb{R}^n$ formulates the linear objective function, $\mathbf{A} \in \mathbb{R}^{m \times n}$ and $\mathbf{b} \in \mathbb{R}^{m}$ formulate the set of $m$ constraints, with $\mathbf{A}$ representing the constraint matrix and $\mathbf{b}$ the vector of constraint bounds, $x_j$ is the $j$-th component of vector $\mathbf{x}$, $I \subseteq \{1, \dots, n\}$ is the set of indices for integer-constrained variables, and $z^*$ represents the optimal value of the objective function in Eq~\ref{def_milp}. 

\textbf{Cutting planes.} Consider the MIP problem defined in Eq~\ref{def_milp}. By relaxing the integer constraints, we obtain the linear programming (LP) relaxation, expressed as:
\begin{equation}
z^*_{LP} = \min_{\mathbf{x}} \left\{\mathbf{c}^T\mathbf{x} \mid \mathbf{A}\mathbf{x} \leq \mathbf{b}, \mathbf{x} \in \mathbb{R}^n \right\}
\label{def_mlp}
\end{equation}

Since this relaxation removes the integer constraints on $\mathbf{x}$, it expands the feasible region of the original MIP problem in Eq~\ref{def_milp}, leading to $z^*_{LP} \leq z^*$. The LP relaxation provides a dual bound, serving as a lower bound for the original MIP. To tighten the LP relaxation while preserving all integer feasible solutions of the original problem, cutting planes (cuts) are introduced. These cuts are linear inequalities added to the LP relaxation to refine its feasible region. B\&C algorithm generates cuts in iterative rounds. In each round $k$, the process includes: (i) solving the current LP relaxation, (ii) generating a set of candidate cuts $C^k$, (iii) selecting a subset $S^k \subseteq C^k$, (iv) adding $S^k$ to the LP relaxation to form a new, tighter relaxation, and (v) moving to the next round. Although adding all generated cuts would ideally tighten the LP relaxation and enhance the dual bound, it can also lead to significantly larger and more complex models. This increase in model size can cause higher computational demands and potential numerical issues~\cite{wesselmann2012implementing,paulus2022learning}. Therefore, selecting an appropriate subset of cuts is essential for maintaining solver efficiency, as adding irrelevant or excessive cuts can lead to diminishing returns and even degrade solver performance.



\section{Motivation}\label{motivation}

\textbf{Inefficiency of root-only cuts.} Restricting cutting planes to the root node is a common configuration in MIP solvers, but this approach is inadequate for large-scale problems involving thousands of nodes. As the solution process progresses, many new cutting planes are generated at subsequent nodes. Without adding new cutting planes at these later nodes, it becomes difficult to make sufficient reductions, significantly slowing down the solving process. Table~\ref{tab:root_cp_experiment} demonstrates significant performance degradation when cutting planes are restricted to the root node, particularly as the problem size increases.
\begin{table}[t]
    \centering\small
    \caption{Performance Comparison of Cutting Plane Strategies}
    \begin{tabular}{lccc}
        \toprule
        Dataset & Strategy & Time (s) & Nodes Explored \\
        \midrule
        \multirow{2}{*}{FCMCNF (p=16)} & only root& 7.72 & 184.3 \\
                                   & all nodes      & 3.55  &  79.8 \\
        \midrule
        \multirow{2}{*}{FCMCNF (p=21)} & only root & 82.08 & 2211.7 \\
                                & all nodes      & 50.83 & 333.5 \\
        \midrule
        \multirow{2}{*}{FCMCNF (p=26)} & only root & 367.81 & 4506.9 \\
                                   & all nodes      & 224.91 & 589.7 \\
        \bottomrule
    \end{tabular}
    \label{tab:root_cp_experiment}
\end{table}

\begin{figure}[t]\centering
	\includegraphics[width=.47\textwidth]{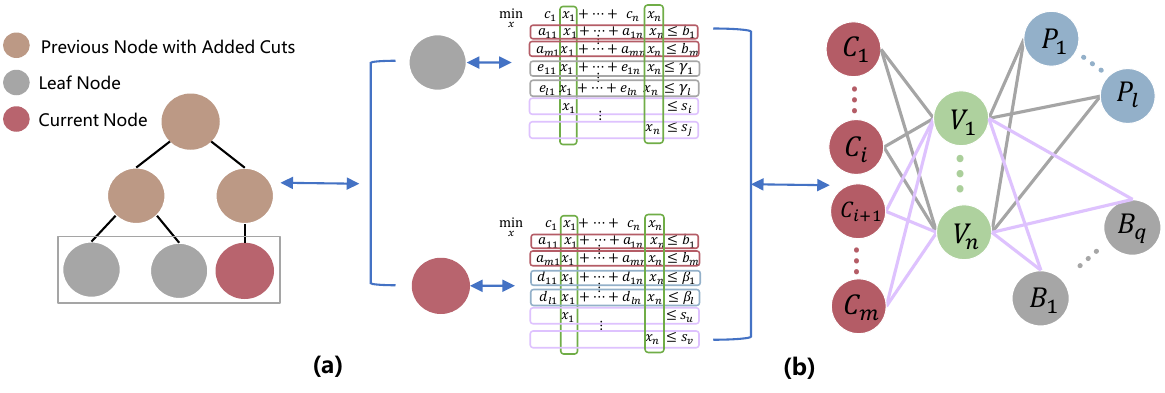}
	\caption{Bipartite graph based on B\&C tree and generated cuts. \textbf{(a) Using all leaf nodes to represent the search tree.} \textbf{(b) Using a bipartite graph to represent all leaf nodes.}}
    \label{embedding}
\end{figure}

\textbf{Lack of context.} Previous learning-based approaches do not leverage the full context of the search tree, only using information from the current node. Their feature representations omit three key aspects: (1) structural variations between nodes (e.g., branching constraints, bound changes), (2) long-term effects of historical cut additions, and (3) coordination between cuts applied at different tree depths. This lack of contextual information can hinder the solver's ability to navigate the search space effectively, particularly for complex, large-scale problems.

\textbf{Challenges in all-node cuts.} Implementing a strategy that adds cutting planes at all nodes presents several challenges. Firstly, it is essential to refine the representation of the entire search tree. For problems of a certain scale, the search tree can be extremely large, often containing a lot of redundant information. Reducing this redundancy and minimizing the data required to represent the entire search tree is crucial. Secondly, designing learning strategies and network architectures for cut plane selection is challenging. A key difficulty arises from the need to develop selection strategies that can fully capture and leverage the intricate structure and context of the search tree, which is often large, sparse, and dynamically evolving. Consequently, a more sophisticated approach is required to make optimal decisions that take into account the global context of the problem and the interdependencies between various components of the optimization model.
\section{Global Cut Selection}

\begin{figure*}[t]\centering 
	\includegraphics[width=\textwidth]{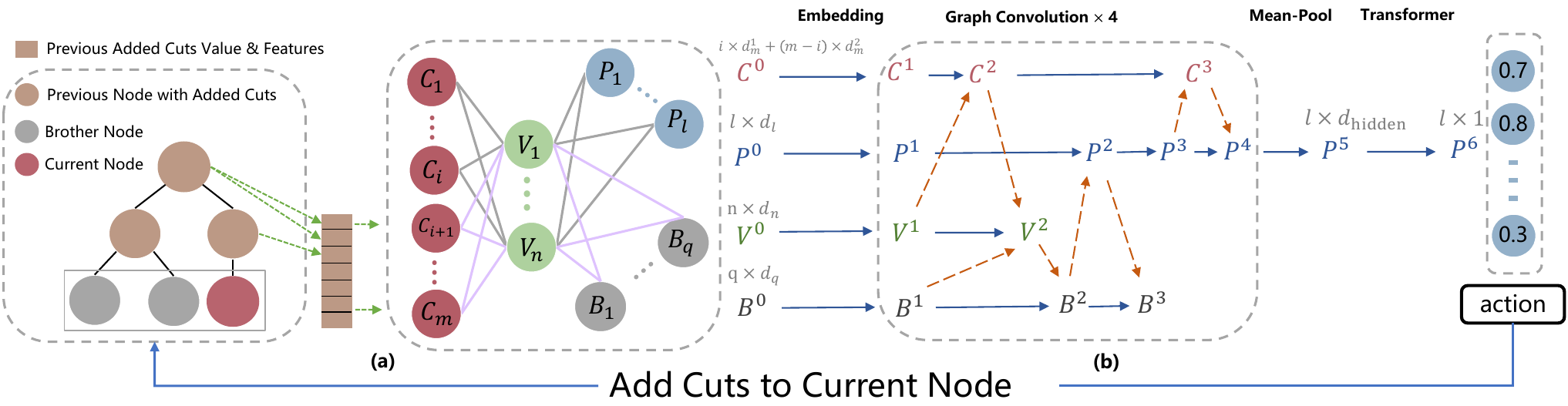}
	\caption{Procedures for adding cutting planes to the current node. \textbf{(a) Bipartite graph representation.} This panel illustrates the encoding of the B\&C tree as a bipartite graph. \textbf{(b) Neural architecture.} This panel outlines the architecture of the neural network.}
    \label{fig:struct}
\end{figure*}

In the task of cut selection, the optimal subset to choose is not directly accessible. However, the quality of the selected subset can be evaluated using a solver, which provides feedback to the learning algorithm. Therefore, we leverage RL to develop cut selection strategies. In this section, we will explain our methodological framework in detail. First, we propose a novel feature extraction method that represents the B\&C tree and the candidate cutting planes as a bipartite graph. We then model the cut selection step in the branch-and-cut algorithm as a Markov Decision Process (MDP). Finally, we explaining the methods used for training the policy network and the rationale behind its architecture and configuration. The overall workflow of the GCS method is shown in Figure~\ref{fig:struct}. A list of the symbols used in this chapter and their corresponding explanations can be found in Appendix~\ref{appendix:symbols}.

\subsection{Constructing Graph from the B\&C Tree and Generated Cuts}\label{graph_rep}

We extend the work of Gasse et al.~\shortcite{gasse2019exact}, who represent MIP as a bipartite graph, and the approach of Paulus et al.~\shortcite{paulus2022learning}, who model a MIP with generated cuts as a tripartite graph. Our approach represents the search tree as a bipartite graph. For each cut selection decision at node, we extract information from the solver to describe the bipartite graph.

Let $T$ represent the branch-and-cut tree, with each node $v \in T$ corresponding to a state in the search process. A node in $T$ is generated by adding cutting plane constraints and branching constraints to its parent node, leading to a significant amount of redundant information across the tree. This redundancy offers an opportunity for optimization in our graph-based representation. As demonstrated by Zhang et al.~\shortcite{zhanglearning}, a parent node $v_p$ can be reconstructed from its child nodes $v_1, v_2 \in T$, allowing us to represent the entire search tree by embedding the leaf nodes $L$.

Each leaf node $v_l \in L$ consists of four key components: the objective function, original problem constraints, cutting plane constraints, and branching constraints. Specifically, at leaf node $v_l$, we have:\\
original objective function:
\begin{equation}
    \min_{x} c_1 x_1 + \dots + c_n x_n \tag{a}
\end{equation}
subject to the original constraints:
\begin{equation} \label{eq:b}
    \sum_{j=1}^{n} a_{ij} x_j \leq b_i, \quad i = 1, \dots, m \tag{b}
\end{equation}
previously added cutting planes:
\begin{equation}\label{eq:c}
    \sum_{j=1}^{n} e_{ij} x_j \leq \gamma_i, \quad i = 1, \dots, q \tag{c}
\end{equation}
and branching constraints:
\begin{equation}
    x_i \leq s_j, \quad \text{or} \quad x_i \geq t_j. \tag{d}
\end{equation}
The bipartite graph encoding captures the essential components of the MIP. Variable nodes $V = \{v_1, \dots, v_n\}$ represent the decision variables, while constraint nodes $C = \{C_1, \dots, C_m\}$ correspond to the original constraints (Fig~\ref{embedding}). Each variable node is connected to the constraint nodes via edges that represent the coefficients $a_{ij}$ in the original constraints. 

In addition to the original constraints, we introduce cutting plane constraints, represented by nodes $B = \{B_1, \dots, B_q\}$ and $C' = \{C_{m+1}, \dots, C_{m+q}\}$, where each cutting plane is characterized by its addition time, improvement effect, and alignment with the objective function (Fig~\ref{embedding}). The branching constraints are encoded as edges with three-dimensional vectors $(x_i, \leq \text{or} >, \text{rhs})$, indicating the branching decision at each node. Candidate cutting planes are also represented as nodes $P = \{P_1, \dots, P_l\}$, summarized by their relationship to the objective function and the current LP solution.

The graph's edge features capture two distinct types of variable-constraint relationships through bipartite connections. For basic constraint-variable edges, each edge $(x_j, C_i)$ inherits the scalar coefficient $a_{ij}$ from the original constraints in Eq~\ref{eq:b}, directly encoding the problem's fundamental structure. Cut-variable edges extend this representation by augmenting constraint coefficients with temporal branching context. Each edge $(x_j, B_k)$ associated with cutting plane $\sum_{j=1}^n e_{kj}x_j \leq \gamma_k$ from Eq~\ref{eq:c} carries a pair(sign,rhs), enabling precise temporal localization of cut additions within the tree structure. For a detailed description of each graph feature, please refer to Appendix~\ref{appendix:graph_feature}.

\subsection{Reinforcement Learning Formulation}

We define the problem as a MDP described by the tuple $(\mathcal{S}, \mathcal{A}, \mathcal{T}, \mathcal{R})$. Specifically, we define the state space $\mathcal{S}$, the action space $\mathcal{A}$, the transition function $\mathcal{T}: \mathcal{S} \times \mathcal{A} \rightarrow \mathcal{S}'$, and the reward function $\mathcal{R}: \mathcal{S} \times \mathcal{A} \rightarrow \mathcal{R}$.

\textbf{(1) The state space $\mathcal{S}$.} Here, \(\mathcal{S} = G\), where \(G\) represents the bipartite graph introduced in Section~\ref{graph_rep}. We expand the state space to encompass the entire search tree, providing comprehensive information for the cut selection algorithm. Specifically, the initial state \(s_0\) is a graph composed of the root node and candidate cutting planes. As the algorithm progresses, the state at the \(t\)-th iteration, denoted \(s_t\), captures the current structure of the search tree along with the updated information about the cutting planes selected so far.

\textbf{(2) The action space \(\mathcal{A}\).} To define the action space \( \mathcal{A} \) for selecting candidate cutting planes \( c_1, c_2, \ldots, c_n \), we adopt the following formulation in the context of combinatorial optimization:
First, consider each cutting plane \( c_i \) as a binary variable representing its selection status:
\begin{equation*}
y_i = 
\begin{cases} 
1, & \text{if } c_i \text{ is selected}, \\
0, & \text{otherwise}.
\end{cases}
\end{equation*}
The selection of cutting planes can be represented by the tuple \((y_1, y_2, \ldots, y_n)\). Next, we must consider the permutations of the selected cutting planes. As Wang et al.~\shortcite{wang2023learning} noted, the arrangement of cuts can significantly impact the subsequent solving algorithms, such as the simplex method, which is highly sensitive to the order of operations. 
Thus, the action space \( \mathcal{A} \) comprises all possible selections and orderings of the cutting planes as:
\begin{equation*}
\begin{aligned}
\mathcal{A} = \Big\{ (y_1, y_2, \ldots, &y_n, \pi) \mid  \, y_i \in \{0, 1\}, \\
& \pi \text{ is a permutation of } \{ i \mid y_i = 1 \} \Big\}.
\end{aligned}
\end{equation*}
The action space includes all possible selections and orderings of the cutting planes, with a complexity of \(O(n \cdot n!)\). For a detailed proof, see Appendix~\ref{appendix:actionspace}.

\textbf{(3) The transition function \(\mathcal{T}:\mathcal{S} \times \mathcal{A} \rightarrow \mathcal{S}\).}  
The transition function maps the current state and action to the next state after incorporating the selected cuts. Specifically, the next state is not necessarily the immediate child node following the addition of cuts and branching, as the transition depends on the updated state of the search tree.

\textbf{(4) The reward function \(\mathcal{R}\).}  
The reward function evaluates the improvement of our method over the baseline, focusing on both gap and time improvements. Rewards are divided into immediate and terminal rewards: the immediate reward reflects gap improvement, while the terminal reward reflects time improvement over SCIP.

\textbf{Immediate reward:}  
The immediate reward \( r(s_t, a_t) \) after selecting cutting planes is:
\(
r(s_t, a_t) = \Delta \text{gap}_t - \text{baseline}(s_t),
\)
where \(\Delta \text{gap}_t = \text{gap}_{t-1} - \text{gap}_t\), and the gap is defined as:
\[
\text{gap} = \frac{|\text{Best Primal} - \text{Best Feasible}|}{|\text{Best Primal}|}.
\]
The baseline term \(\text{baseline}(s_t)\) is the gap improvement achieved by SCIP at time \(t\).

\textbf{Terminal reward:}  
The terminal reward is the difference in solving times between SCIP and our method:
\[
r(s_T, a_T) = T_{\text{baseline}} - T_{\text{method}},
\]
where \(T_{\text{baseline}}\) and \(T_{\text{method}}\) are the solving times for SCIP and our method, respectively.

\subsection{Policy Network Architecture}

First, we extract features from the bipartite graph nodes, representing the constraints (\(C\)), decision variables (\(V\)), previously added cutting planes (\(B\)), and candidate cutting planes (\(P\)). These features are embedded into uniform dimension vectors using an embedding layer:
\begin{equation*}
   \mathbf{h}_i = \text{Embed}(\mathbf{f}_i), \quad i \in \{C, V, B, P\}
\end{equation*}
Next, we apply four iterations of graph convolution \cite{kipf2016semi,scarselli2008graph} to aggregate information from neighboring nodes in the graph. At each iteration, each node's feature vector is updated based on the features of its neighbors, using a learnable weight matrix and a non-linear activation function. This allows nodes to exchange information, capturing the structure and relationships in the graph. After the graph convolution steps, we apply mean pooling over the candidate cutting plane nodes (\(P\)) to summarize their features:
\begin{equation*}
   \mathbf{h}_P = \frac{1}{l} \sum_{i=1}^{l} \mathbf{h}_i^{(\text{final})}
\end{equation*}
where \( \mathbf{h}_i^{(\text{final})} \) is the feature vector of node \( P_i \) after the final graph convolution.

The pooled features are then passed through a transformer sequence model~\cite{vaswani2017attention}, which effectively evaluates the combined effects of the different cutting planes. The transformer captures the interactions and dependencies between the cutting planes, generating hidden states \( \mathbf{H}_{\text{transformer}} \) as the output.

Finally, the policy network uses a sigmoid activation function to output the probability of selecting each cutting plane \( P_i \). The selection probability is computed as:
\begin{equation*}
   p_i = \sigma\left( \mathbf{h}_{P_i} \cdot W_{\text{policy}} + b_{\text{policy}} \right)
\end{equation*}
where \( \mathbf{h}_{P_i} \) is the feature vector of the cutting plane, \( W_{\text{policy}} \) is the weight matrix, and \( b_{\text{policy}} \) is the bias term. This probability determines which cutting plane will be selected during the optimization process, guiding the solver's decision-making.

\subsection{Training}
We utilize the Proximal Policy Optimization (PPO) algorithm~\cite{schulman2017proximal} to train our decision network, allowing it to effectively explore and learn in high-dimensional action spaces. PPO is an on-policy reinforcement learning algorithm that strikes a balance between ease of implementation and computational efficiency. It alternates between data collection and optimizing a surrogate objective function. This approach ensures that the network explores the environment sufficiently while keeping the updates stable.
Specifically, the PPO algorithm alternates between data collection and optimizing a surrogate objective function. The algorithm uses a clipped objective function for stable training, optimizing the policy and value networks. The policy network updates aim to maximize expected advantages within clipped probability ratios, while the value network minimizes the mean squared error between predicted and actual rewards. For a more detailed description of the training algorithm, please refer to Appendix~\ref{appendix:train}.

\subsection{Discussion on Advantages of GCS}
\textbf{Fast refreshing of bipartite graph.} Before selecting cutting planes in each iteration, we need to embed the current search tree into a bipartite graph. Our graph structure allows for efficient updates, requiring only updates to the cutting plane nodes after each iteration. We achieve this by maintaining a set of all added cutting planes along with their corresponding branching constraints, facilitating rapid reconstruction of the graph. Each time the cutting planes are added or branching occurs, the graph is updated in $O(1)$ time. This updatable graph design allows for the efficient representation of the search tree within the solver. Without the ability to quickly update, representing the search tree for large-scale problems would demand substantial computational time in each iteration.

\textbf{Efficient training with fewer samples.}
GCS can achieve effective training with fewer samples compared to traditional learning-based methods that only add cutting planes at the root node. By selecting cuts at all nodes in the search tree, GCS generates a larger number of state-reward pairs in each iteration. This leads to a more diverse set of training samples, allowing the model to learn from a broader range of states and actions. As a result, the training process becomes more efficient, as the model can converge using fewer problem instances. In contrast, methods that limit cut selection to the root node generate fewer state-reward pairs per iteration, which means that the model requires more problem instances to achieve the same level of training. By utilizing more informative samples from the entire tree, GCS accelerates the learning process and reduces the amount of data needed for training, making it more scalable and efficient in large-scale settings.

\section{Experiments}
\begin{table*}[t]
\centering
\caption{Comparison of the default policy for SCIP, no cuts (NoCuts), score-based policy (SBP), hierarchical sequence model (HEM), and GCS (Ours) across various datasets. The best is marked in bold.}
\label{table:result1} 
\scriptsize
\begin{tabularx}{\textwidth}{l>
{\centering\arraybackslash}X>{\centering\arraybackslash}X>{\centering\arraybackslash}X>
{\centering\arraybackslash}X>{\centering\arraybackslash}X>{\centering\arraybackslash}X>
{\centering\arraybackslash}X>{\centering\arraybackslash}X>{\centering\arraybackslash}X>
{\centering\arraybackslash}X>{\centering\arraybackslash}X>{\centering\arraybackslash}X}
\toprule
\toprule
 & \multicolumn{4}{c}{\textbf{Set Covering}} & \multicolumn{4}{c}{\textbf{GISP}} & \multicolumn{4}{c}{\textbf{MIS}} \\ 
 & \multicolumn{4}{c}{($n = 500$, $m = 1000$)} & \multicolumn{4}{c}{($p = 95$)} & \multicolumn{4}{c}{($n = 1953$, $m = 500$)} \\ 
\midrule
\multirow{2}{*}{\textbf{Methods}} 
& \multirow{2}{*}{\textbf{Time (s) $\downarrow$}} &
\textbf{Improve} \newline \textbf{(Time)$\uparrow$}
& \multirow{2}{*}{\textbf{Nodes$\downarrow$}} &\textbf{Reduction (Node)$\uparrow$}
& \multirow{2}{*}{\textbf{Time (s) $\downarrow$}} &
\textbf{Improve} \newline \textbf{(Time)$\uparrow$}
& \multirow{2}{*}{\textbf{Nodes$\downarrow$}} &\textbf{Reduction (Node)$\uparrow$}
& \multirow{2}{*}{\textbf{Time (s) $\downarrow$}} &
\textbf{Improve} \newline \textbf{(Time)$\uparrow$}
& \multirow{2}{*}{\textbf{Nodes$\downarrow$}} &\textbf{Reduction (Node)$\uparrow$}\\ 
 \cmidrule(r){1-5}  \cmidrule(r){6-9} \cmidrule(r){10-13}
SCIP    & 10.33 & NA & 160.6 & NA
        & 24.58 & NA & 1963.5 & NA
        & 7.27  & NA & 65.1 &  NA\\ 
NoCuts  & 13.85 & -34.1\%& 190.5 & -18.6\%
        & 22.01 & 10.5\% & 13525.7 & -588.9\%
        & 10.08 & -38.7\%& 2499.6 & -3739\%\\ 
SBP     & 9.64   & 6.7\% &  115.6 & 28.0\%
        & 28.51  & -16.0\% & 1847.7 & 5.9 \%
        & 5.74   & 21.0\% & 54.5 & 16.3\%\\ 
HEM     & 8.53  & 17.4\% & \textbf{107.6} & \textbf{33.0}\%
        & 26.62 & -7.7\%& 2366.7 & -20.5\%
        & 5.61  & 22.8 \%& 56.7 & 13.0\%\\ 
\midrule
GCS (Ours) & \textbf{5.79}   & \textbf{43.9\%} & 109.5 & 31.8\%
            & \textbf{18.39} & \textbf{25.2\%} & \textbf{1444.5} & \textbf{26.4\%}
            & \textbf{4.75} & \textbf{34.7\%} & \textbf{53.9}  & \textbf{17.2\%}\\ 
\bottomrule
\\
\toprule
\toprule
 & \multicolumn{4}{c}{\textbf{FCMCNF}} & \multicolumn{4}{c}{\textbf{MIK}} & \multicolumn{4}{c}{\textbf{Corlat}}  \\ 
 & \multicolumn{4}{c}{($p = 21$)} & \multicolumn{4}{c}{($n = 346$, $m = 413$)} & \multicolumn{4}{c}{($n = 486$, $m = 466$)} \\ 
\midrule
\multirow{2}{*}{\textbf{Methods}} 
& \multirow{2}{*}{\textbf{Time (s) $\downarrow$}} &
\textbf{Improve} \newline \textbf{(Time)$\uparrow$}
& \multirow{2}{*}{\textbf{Nodes$\downarrow$}} &\textbf{Reduction (Node)$\uparrow$}
& \multirow{2}{*}{\textbf{Time (s) $\downarrow$}} &
\textbf{Improve} \newline \textbf{(Time)$\uparrow$}
& \multirow{2}{*}{\textbf{Nodes$\downarrow$}} &\textbf{Reduction (Node)$\uparrow$}
& \multirow{2}{*}{\textbf{Time (s) $\downarrow$}} &
\textbf{Improve} \newline \textbf{(Time)$\uparrow$}
& \multirow{2}{*}{\textbf{Nodes$\downarrow$}} &\textbf{Reduction (Node)$\uparrow$}\\ 
 \cmidrule(r){1-5}  \cmidrule(r){6-9} \cmidrule(r){10-13}
SCIP & 50.83  & NA & 333.5 & NA
    & 17.06  & NA & 2573.9 & NA
    &  30.36  & NA & 5318.8 & NA\\  
NoCuts & 1800* & -3441\%* & 48608* & {\tiny -14475\%*}
        &  354.92   & -1980\%  & 397210 & -15332\%
        &  1800*  & -5828\%*  & 721251* & {\tiny -13460\%*}\\ 
SBP & 49.85 & 1.9\% & 297.1 & 10.9\%
    & 39.80  & -133.3\% & 25447.4 & -888.7\%
    & 26.15  & 13.9\% & 5668.3 & -6.6\%\\ 
HEM & 48.94 & 3.7\% & 350.7 & 5.2\%
    & 23.48 & -37.6\% & \textbf{2392.9} & \textbf{7.0\%}
    & 24.54 & 19.2\% & \textbf{4381.4} & \textbf{17.6\%}\\
\midrule
GCS (Ours) & \textbf{37.18} & \textbf{26.9\%} & \textbf{208.4} & \textbf{37.5\%}
            & \textbf{16.91}  & \textbf{0.9\%}  & 3012.0 & -17.0\%
            & \textbf{24.17}  & \textbf{20.4\%}  & 4892.5 & 8.0\%\\ 
\bottomrule
\end{tabularx}
\end{table*}

\begin{table*}[t]
\centering
\begin{minipage}{\textwidth}
\centering
\caption{Ablation experiments comparing the performance of the SCIP default policy, HEM, and GCS when cutting planes are added exclusively at the root node versus at all nodes within the B\&C tree.}
\label{table:result3} 
\scriptsize
\begin{tabularx}{\textwidth}{l>
{\centering\arraybackslash}X>{\centering\arraybackslash}X>{\centering\arraybackslash}X>
{\centering\arraybackslash}X>{\centering\arraybackslash}X>{\centering\arraybackslash}X>
{\centering\arraybackslash}X>{\centering\arraybackslash}X>{\centering\arraybackslash}X>
{\centering\arraybackslash}X>{\centering\arraybackslash}X>{\centering\arraybackslash}X}
\toprule
\toprule
 & \multicolumn{4}{c}{\textbf{Set Covering}} & \multicolumn{4}{c}{\textbf{GISP}} & \multicolumn{4}{c}{\textbf{MIS}} \\ 
 & \multicolumn{4}{c}{($n = 500$, $m = 1000$)} & \multicolumn{4}{c}{($p = 95$)} & \multicolumn{4}{c}{($n = 1953$, $m = 500$)} \\ 
\midrule
\multirow{2}{*}{\textbf{Methods}} 
& \multirow{2}{*}{\textbf{Time (s) $\downarrow$}} &
\textbf{Improve} \newline \textbf{(Time)$\uparrow$}
& \multirow{2}{*}{\textbf{Nodes$\downarrow$}} &\textbf{Reduction (Node)$\uparrow$}
& \multirow{2}{*}{\textbf{Time (s) $\downarrow$}} &
\textbf{Improve} \newline \textbf{(Time)$\uparrow$}
& \multirow{2}{*}{\textbf{Nodes$\downarrow$}} &\textbf{Reduction (Node)$\uparrow$}
& \multirow{2}{*}{\textbf{Time (s) $\downarrow$}} &
\textbf{Improve} \newline \textbf{(Time)$\uparrow$}
& \multirow{2}{*}{\textbf{Nodes$\downarrow$}} &\textbf{Reduction (Node)$\uparrow$}\\ 
 \cmidrule(r){1-5}  \cmidrule(r){6-9} \cmidrule(r){10-13}
SCIP(only root) & 13.36 & -29.3\%& 183.5 & -14.4\%
                & 30.71 & -24.9\% & 3539.4 & -80.3\%
                & 7.87  & -8.3\% & 106.1 & -63.0    \\  
SCIP(all nodes)  & 10.33 & NA & 160.6 & NA
                & 24.58 & NA & 1963.5 & NA
                & 7.27  & NA & 65.1 &  NA\\
HEM(only root)  & 9.53  & 7.7\%  & 125.1    & 22.1\%
                & 28.84 &  -17.3\% & 2581.9    & -31.5\%
                & 6.60  & 9.2\%  & 59.7      & 8.3\%\\
HEM(all nodes)   & 8.53  & 17.4\% & \textbf{107.6} & \textbf{33.0}\%
                & 26.62 & -7.7\%& 2366.7 & -20.5\%
                & 5.61  & 22.8 \%& 56.7 & 13.0\%\\ 
\midrule
GCS(only root) & 9.37  & 9.3\%   & 141.3 & 12.0\%
                & 30.11  & -22.5\% & 2956.0& - 50.5\%
                & 7.71  & -6.1\% & 87.0 & -33.6\%    \\ 
GCS(all nodes) & \textbf{5.79}   & \textbf{43.9\%} & 109.5 & 31.8\%
                & \textbf{18.39} & \textbf{25.2\%} & \textbf{1444.5} & \textbf{26.4\%}
                & \textbf{4.75} & \textbf{34.7\%} & \textbf{53.9}  & \textbf{17.2\%}\\ 
\bottomrule
\\

\toprule
\toprule
 & \multicolumn{4}{c}{\textbf{FCMCNF}} & \multicolumn{4}{c}{\textbf{MIK}} & \multicolumn{4}{c}{\textbf{Corlat}}  \\ 
 & \multicolumn{4}{c}{($p = 21$)} & \multicolumn{4}{c}{($n = 346$, $m = 413$)} & \multicolumn{4}{c}{($n = 486$, $m = 466$)} \\ 
\midrule
\multirow{2}{*}{\textbf{Methods}} 
& \multirow{2}{*}{\textbf{Time (s) $\downarrow$}} &
\textbf{Improve} \newline \textbf{(Time)$\uparrow$}
& \multirow{2}{*}{\textbf{Nodes$\downarrow$}} &\textbf{Reduction (Node)$\uparrow$}
& \multirow{2}{*}{\textbf{Time (s) $\downarrow$}} &
\textbf{Improve} \newline \textbf{(Time)$\uparrow$}
& \multirow{2}{*}{\textbf{Nodes$\downarrow$}} &\textbf{Reduction (Node)$\uparrow$}
& \multirow{2}{*}{\textbf{Time (s) $\downarrow$}} &
\textbf{Improve} \newline \textbf{(Time)$\uparrow$}
& \multirow{2}{*}{\textbf{Nodes$\downarrow$}} &\textbf{Reduction (Node)$\uparrow$}\\ 
 \cmidrule(r){1-5}  \cmidrule(r){6-9} \cmidrule(r){10-13}
SCIP(only root) & 82.08  & -61.5\%  & 2211.7     & -563.2\%
                & 27.76  & -62.7\%  & 16109.4    & -525.9\%
                & 68.04  & -124.1\% & 34356.0    & -545.9\%  \\  
SCIP(all nodes)  & 50.83  & NA & 333.5 & NA
                & 17.06  & NA & 2573.9 & NA
                &  30.36  & NA & 5318.8 & NA\\   
HEM(only root)  & 68.65 & -35.1\%  & 769.2 & -130.6\%
                & 31.57 & -85.1\%  & 11502.1& -346.9\%
                & 52.42 & -72.7\%  & 10039.7 & -88.8\%\\
HEM(all nodes)   & 48.94 & 3.7\% & 350.7 & -5.2\%
                & 23.48 & -37.6\% & \textbf{2392.9} & \textbf{7.0\%}
                & 24.54 & 19.2\% & \textbf{4381.4} & \textbf{17.6\%}\\
\midrule
GCS(only root) & 65.88  &  -29.6\%   & 724.0 & -117.1\%
                 & 33.96  &  -99.1\%   & 22126.0 & -759.6\%
                 & 54.14  &  -78.3\%   & 20191.3  & -279.6\%\\ 
GCS(all nodes) & \textbf{37.18} & \textbf{26.9\%} & \textbf{208.4} & \textbf{37.5\%}
                & \textbf{16.91}  & \textbf{0.9\%}  & 3012.0 & -17.0\%
                & \textbf{24.17}  & \textbf{20.4\%}  & 4892.5 & 8.0\%\\ 
\bottomrule

\end{tabularx}
\end{minipage}
\end{table*}

\begin{table*}[t]
\centering
\begin{minipage}{\textwidth}
\centering
\caption{Generalization tests evaluating the performance of GCS on the FCMCNF problem. The method is trained on instances with \(p = 13\) and tested on larger scales with \(p = 16\), \(p = 21\), and \(p = 26\).}
\label{table:result2} 
\scriptsize
\begin{tabularx}{\textwidth}{l>
{\centering\arraybackslash}X>{\centering\arraybackslash}X>{\centering\arraybackslash}X>
{\centering\arraybackslash}X>{\centering\arraybackslash}X>{\centering\arraybackslash}X>
{\centering\arraybackslash}X>{\centering\arraybackslash}X>{\centering\arraybackslash}X>
{\centering\arraybackslash}X>{\centering\arraybackslash}X>{\centering\arraybackslash}X}
\toprule
\midrule
& \multicolumn{3}{c}{\textbf{FCMCNF}} & \multicolumn{3}{c}{\textbf{FCMCNF}} & \multicolumn{3}{c}{\textbf{FCMCNF}} & \multicolumn{3}{c}{\textbf{FCMCNF}} \\ 
& \multicolumn{3}{c}{($p = 13$)} & \multicolumn{3}{c}{($p = 16$)} & \multicolumn{3}{c}{($p = 21$)} & \multicolumn{3}{c}{($p = 26$)} \\ 
\midrule
\multirow{2}{*}{\textbf{Methods}} 
& \multirow{2}{*}{\textbf{Time (s) $\downarrow$}} &
\textbf{Improve} \newline \textbf{(Time)$\uparrow$}
& \multirow{2}{*}{\textbf{Nodes$\downarrow$}} 
& \multirow{2}{*}{\textbf{Time (s) $\downarrow$}} &
\textbf{Improve} \newline \textbf{(Time)$\uparrow$}
& \multirow{2}{*}{\textbf{Nodes$\downarrow$}} 
& \multirow{2}{*}{\textbf{Time (s) $\downarrow$}} &
\textbf{Improve} \newline \textbf{(Time)$\uparrow$}
& \multirow{2}{*}{\textbf{Nodes$\downarrow$}} 
& \multirow{2}{*}{\textbf{Time (s) $\downarrow$}} &
\textbf{Improve} \newline \textbf{(Time)$\uparrow$}
& \multirow{2}{*}{\textbf{Nodes$\downarrow$}} \\ 
\cmidrule(r){1-4}  \cmidrule(r){5-7} \cmidrule(r){8-10} \cmidrule{11-13}
SCIP    & 2.40   & NA & 64.0
        & 3.55   & NA & 79.8
        & 50.83  & NA & 333.5
        & 224.91 & NA & 589.7 \\ 
\midrule
SBP     & 1.96  & 18.3\% & \textbf{50.2}
        & 3.18  & 10.4\%  & 65.3
        & 50.62 & 0.4\%  & 427.0
        & 268.14 & -19.2\% & 754.5\\ 
HEM & 1.86  &  22.5\%    & 59.8
    & 2.90  & 18.3\% & \textbf{54.9}
    & 51.18 & -6.9\% & \textbf{331.2}
    & 261.75  & -16.4\%  & \textbf{516.8}\\ 
\midrule
GCS (Ours) & \textbf{1.85}     & \textbf{22.9\%}    & 58.6
            & \textbf{2.82}      &  \textbf{20.6\%}    & 64.7
            & \textbf{42.99}     & \textbf{15.4\%}    & 389.3
            & \textbf{205.03}    & \textbf{8.8\%}     & 575.9 \\
\bottomrule
\end{tabularx}
\end{minipage}
\end{table*}

We divide our experimental section into three parts. In \textbf{Experiment 1}, we evaluate our method on six classical MIP datasets from various application domains. In \textbf{Experiment 2}, we conduct an ablation study comparing the performance of the SCIP default method and the AI method when adding cutting planes only at the root node versus adding them at all nodes. \textbf{Experiment 3} focuses on the generated dataset FCMCNF, testing whether the AI method can generalize to instances significantly larger than those seen during training.

\subsection{Experimental Setup}
We use SCIP 8.0.4~\cite{scip} as the backend solver throughout all experiments. SCIP is a state-of-the-art open-source solver widely adopted in research related to machine learning for combinatorial optimization~\cite{chmiela2021learning,gasse2019exact,turner2022adaptive,wang2023learning}. To ensure the fairness and reproducibility of our comparisons, we retain all default parameters in SCIP. In contrast to previous work \cite{wang2023learning}, which limited cut generation and selection to the root node, our setup allows the solver to generate and select cuts at all nodes, leading to enhanced performance. Additionally, previous work set SCIP's hyperparameters to allow separators to generate more cuts, thereby highlighting the effectiveness of single-node selection strategies. However, attempting to generate more cuts at all nodes with maximum effort can lead to significant performance loss. Therefore, unlike prior studies that maximized the frequency of cut separators, we maintain all default SCIP hyperparameters. The model is implemented in PyTorch~\cite{paszke2019pytorch} and optimized using the Adam optimizer~\cite{kingma2014adam}. The training process is conducted on a single machine equipped with eight NVIDIA GeForce RTX 4090 GPUs and two AMD EPYC 7763 CPUs.

\textbf{Benchmarks.} 
We focus on problems that require solving within a branch-and-cut framework, rather than those that can be solved at the root node alone. We evaluated our method on datasets from six NP-hard problems, including four classical synthetic MIP problems and two real-world MIP problems from diverse application areas. (1) The four classical synthetic MIP problems include: Set Covering~\cite{balas1980set}, Generalized Independent Set (GISP)~\cite{colombi2017generalized}, Maximum Independent Set (MIS)~\cite{bergman2016Decision} and Fixed Charge Multicommodity Network Flow (FCMCNF)~\cite{hewitt2010combining}. We generate instances based on the methods described by \cite{bejar2009generating,chmiela2021learning,labassi2022learning}. (2) Real-world datasets comprise MIK~\cite{atamturk2003facets} and CORLAT~\cite{gomes2008connections}, which are widely used benchmarks for evaluating MIP solvers ~\cite{he2014learning,nair2020solving}. In Table \ref{table:result1},\ref{table:result3},\ref{table:result2}, we indicate the size of each dataset. Here, $n$ and $m$ represent the number of variables and the number of constraints in the dataset, respectively. $p$ denotes the number of vertices in the randomly generated graph. The FCMCNF and GISP datasets are MIP problems derived from this graph. For comprehensive dataset specifications, see Appendix~\ref{appendix:dataset_details} and Appendix~\ref{appendix:dataset_stat}

\textbf{Baselines.} Our baseline methods comprise two hard-coded cut selection strategies and two advanced learning-based methods, including the current SOTA method. The hard-coded strategies include NoCuts and Default. NoCuts signifies that no cuts are added during the solving process, while Default refers to the standard cut selection strategy employed by SCIP 8.0.4~\cite{scip}. Following Wang et al.~\shortcite{wang2023learning}, we use a slight variant of the learning-based methods~\cite{tang2020reinforcement,huang2022learning}, specifically the score-based policy (SBP), as one of our baselines. Furthermore, we evaluate the performance of the SOTA method known as HEM \cite{wang2023learning}.

\textbf{Metrics.} We employ well-established evaluation metrics to assess the performance of our proposed cutting plane selection strategy: (1) the average number of nodes processed per instance during the solving process, and (2) the average runtime in seconds per instance. Additionally, we introduce an \textit{Improvement} metric to compare the performance of different cutting plane selection methods against the default SCIP solver. This metric is defined as
\(
\mathrm{Im}(M) = \frac{T(\mathrm{SCIP}) - T(M)}{T(\mathrm{SCIP})},
\)
where $T(\mathrm{SCIP})$ is the solving time of SCIP, and $T(M)$ is the solving time of the method under comparison. The improvement metric quantifies the relative performance enhancement of a given method compared to SCIP. Furthermore, the total number of nodes processed is also included as a key metric, offering insight into the efficiency of the cutting plane selection strategy. The reduction in the number of nodes processed is analyzed to assess the ability of our methods to simplify the problem space. Both time and node count are critical for evaluating solver performance, as they reflect not only the speed of the solution process but also the computational resources required. Collectively, these metrics provide a comprehensive evaluation of the effectiveness of our proposed strategies. In our experiments, we set the solving time limit to 1800 seconds. For datasets where most instances were challenging to solve within this time limit, we annotate these cases with an asterisk (*) next to the corresponding data.

\subsection{Experiment Results}


\textbf{Experiment (1): Comparative Evaluation.} The results presented in Table~\ref{table:result1} demonstrate that GCS consistently outperforms all baseline methods across the tested datasets. Compared to the default SCIP strategy, GCS significantly reduces solving time and the total number of nodes processed. Specifically, in the Set Covering dataset, GCS reduces solving time by up to 43.9\% and the number of nodes by 31.8\%. Similar improvements are observed in the FCMCNF and MIK datasets, where GCS shows notable efficiency gains.

While the score-based policy (SBP) provides improvements on several datasets, its reliance on predefined scoring mechanisms leads to suboptimal cut selection, as seen in the GISP dataset, where its performance drops by 16.0\% compared to SCIP. Additionally, HEM achieves good results in many cases but shows performance variability on medium and hard instances, indicating that its cut selection model may struggle to adapt across diverse problem instances. We also tested our method on the MIPLIB 2017 dataset, as examples of datasets suitable for cut selection tasks are limited. The results for each test case are provided in Appendix~\ref{appendix:miplib}.

\textbf{Experiment (2): Ablation Study.} This experiment investigates the impact of adding cutting planes exclusively at the root node versus all nodes within the Branch-and-Cut tree. The results in Table~\ref{table:result3} reveal that restricting cuts to the root node leads to significantly worse performance. For example, in the GISP dataset, SCIP with cuts only at the root node shows a 24.9\% degradation in solving time compared to the approach that uses cuts at all nodes. GCS shows a 43.9\% improvement in solving time when cuts are added at all nodes, emphasizing the importance of utilizing information from the entire search tree. Similarly, HEM also benefits from applying cuts at all nodes, demonstrating a clear advantage over restricting cuts to the root node.

\textbf{Experiment (3): Generality Tests.} In this experiment, we evaluate the ability of GCS to generalize across larger MIP instances. Training GCS on smaller instances of the FCMCNF problem (with \(p = 13\) ) and and testing it on larger instances (with \(p = 16\), \(p = 21\), and \(p = 26\)) demonstrates that GCS generalizes well to these larger problems. As shown in Table~\ref{table:result2}, GCS consistently outperforms the baselines, confirming its robustness across a variety of problem scales. This ability to scale efficiently to larger instances is crucial for real-world applications, where the size and complexity of MIP problems can vary significantly. Both SBP and HEM maintain competitive performance, further validating their utility as baselines. 


\section{Conclusion}

Although previous research has shown the effectiveness of reinforcement learning for cut selection at individual nodes, this advantage typically applies to small-scale problems and specific hyperparameters. We have extended this strategy to select cutting planes across all nodes, utilizing information from the entire search tree. However, large instances with oversized search trees may hinder feature extraction, leading to near-optimal results. We believe the proposed GCS can be enhanced by using smaller-scale inputs to speed up feature extraction and designing the guidance network to better determine when to add cutting planes, reducing both training and solving burdens. This approach would improve efficiency and increase adaptability to diverse problem sizes and complexities.


\bibliographystyle{named}
\bibliography{ijcai25}

\begin{thebibliography}{}

\bibitem[\protect\citeauthoryear{Achterberg}{2007}]{achterberg2007constraint}
Tobias Achterberg.
\newblock {\em Constraint integer programming}.
\newblock PhD thesis, Berlin Institute of Technology, 2007.

\bibitem[\protect\citeauthoryear{Atamt{\"u}rk}{2003}]{atamturk2003facets}
Alper Atamt{\"u}rk.
\newblock On the facets of the mixed--integer knapsack polyhedron.
\newblock {\em Mathematical Programming}, 98(1-3):145--175, 2003.

\bibitem[\protect\citeauthoryear{Balas and Ho}{1980}]{balas1980set}
Egon Balas and Andrew Ho.
\newblock {\em Set covering algorithms using cutting planes, heuristics, and
  subgradient optimization: a computational study}.
\newblock Springer, 1980.

\bibitem[\protect\citeauthoryear{Balcan \bgroup \em et al.\egroup
  }{2021}]{balcan2021sample}
Maria-Florina~F Balcan, Siddharth Prasad, Tuomas Sandholm, and Ellen Vitercik.
\newblock Sample complexity of tree search configuration: Cutting planes and
  beyond.
\newblock {\em Advances in Neural Information Processing Systems},
  34:4015--4027, 2021.

\bibitem[\protect\citeauthoryear{Balcan \bgroup \em et al.\egroup
  }{2022}]{balcan2022structural}
Maria-Florina~F Balcan, Siddharth Prasad, Tuomas Sandholm, and Ellen Vitercik.
\newblock Structural analysis of branch-and-cut and the learnability of gomory
  mixed integer cuts.
\newblock {\em Advances in Neural Information Processing Systems},
  35:33890--33903, 2022.

\bibitem[\protect\citeauthoryear{Baltean-Lugojan \bgroup \em et al.\egroup
  }{2019}]{baltean2019scoring}
Radu Baltean-Lugojan, Pierre Bonami, Ruth Misener, and Andrea Tramontani.
\newblock Scoring positive semidefinite cutting planes for quadratic
  optimization via trained neural networks.
\newblock {\em preprint: http://www. optimization-online. org/DB\_
  HTML/2018/11/6943. html}, 2019.

\bibitem[\protect\citeauthoryear{B{\'e}jar \bgroup \em et al.\egroup
  }{2009}]{bejar2009generating}
Ram{\'o}n B{\'e}jar, Alba Cabiscol, Felip Many{\`a}, and Jordi Planes.
\newblock Generating hard instances for maxsat.
\newblock In {\em 2009 39th International Symposium on Multiple-Valued Logic},
  pages 191--195. IEEE, 2009.

\bibitem[\protect\citeauthoryear{Bergman \bgroup \em et al.\egroup
  }{2016}]{bergman2016Decision}
David Bergman, Andre~A Cire, Willem-Jan Van~Hoeve, and John Hooker.
\newblock {\em Decision diagrams for optimization}, volume~1.
\newblock Springer, 2016.

\bibitem[\protect\citeauthoryear{Berthold \bgroup \em et al.\egroup
  }{2022}]{berthold2022learning}
Timo Berthold, Matteo Francobaldi, and Gregor Hendel.
\newblock Learning to use local cuts.
\newblock {\em arXiv preprint arXiv:2206.11618}, 2022.

\bibitem[\protect\citeauthoryear{Bestuzheva \bgroup \em et al.\egroup
  }{2021a}]{bestuzheva2021scip}
Ksenia Bestuzheva, Mathieu Besan{\c{c}}on, Wei-Kun Chen, Antonia Chmiela, Tim
  Donkiewicz, Jasper van Doornmalen, Leon Eifler, Oliver Gaul, Gerald Gamrath,
  Ambros Gleixner, et~al.
\newblock The scip optimization suite 8.0.
\newblock {\em arXiv preprint arXiv:2112.08872}, 2021.

\bibitem[\protect\citeauthoryear{Bestuzheva \bgroup \em et al.\egroup
  }{2021b}]{scip}
Ksenia Bestuzheva, Mathieu Besan{\c{c}}on, Wei-Kun Chen, Antonia Chmiela, Tim
  Donkiewicz, Jasper van Doornmalen, Leon Eifler, Oliver Gaul, Gerald Gamrath,
  Ambros Gleixner, Leona Gottwald, Christoph Graczyk, Katrin Halbig, Alexander
  Hoen, Christopher Hojny, Rolf van~der Hulst, Thorsten Koch, Marco
  L{\"u}bbecke, Stephen~J. Maher, Frederic Matter, Erik M{\"u}hmer, Benjamin
  M{\"u}ller, Marc~E. Pfetsch, Daniel Rehfeldt, Steffan Schlein, Franziska
  Schl{\"o}sser, Felipe Serrano, Yuji Shinano, Boro Sofranac, Mark Turner,
  Stefan Vigerske, Fabian Wegscheider, Philipp Wellner, Dieter Weninger, and
  Jakob Witzig.
\newblock {The SCIP Optimization Suite 8.0}.
\newblock Technical report, Optimization Online, December 2021.

\bibitem[\protect\citeauthoryear{Bixby}{1992}]{bixby1992implementing}
Robert~E Bixby.
\newblock Implementing the simplex method: The initial basis.
\newblock {\em ORSA Journal on Computing}, 4(3):267--284, 1992.

\bibitem[\protect\citeauthoryear{Chen}{2010}]{chen2010integrated}
Zhi-Long Chen.
\newblock Integrated production and outbound distribution scheduling: review
  and extensions.
\newblock {\em Operations research}, 58(1):130--148, 2010.

\bibitem[\protect\citeauthoryear{Chmiela \bgroup \em et al.\egroup
  }{2021}]{chmiela2021learning}
Antonia Chmiela, Elias Khalil, Ambros Gleixner, Andrea Lodi, and Sebastian
  Pokutta.
\newblock Learning to schedule heuristics in branch and bound.
\newblock {\em Advances in Neural Information Processing Systems},
  34:24235--24246, 2021.

\bibitem[\protect\citeauthoryear{Colombi \bgroup \em et al.\egroup
  }{2017}]{colombi2017generalized}
Marco Colombi, Renata Mansini, and Martin Savelsbergh.
\newblock The generalized independent set problem: Polyhedral analysis and
  solution approaches.
\newblock {\em European Journal of Operational Research}, 260(1):41--55, 2017.

\bibitem[\protect\citeauthoryear{Gasse \bgroup \em et al.\egroup
  }{2019}]{gasse2019exact}
Maxime Gasse, Didier Ch{\'e}telat, Nicola Ferroni, Laurent Charlin, and Andrea
  Lodi.
\newblock Exact combinatorial optimization with graph convolutional neural
  networks.
\newblock {\em Advances in neural information processing systems}, 32, 2019.

\bibitem[\protect\citeauthoryear{Gleixner \bgroup \em et al.\egroup
  }{2021}]{gleixner2021miplib}
Ambros Gleixner, Gregor Hendel, Gerald Gamrath, Tobias Achterberg, Michael
  Bastubbe, Timo Berthold, Philipp Christophel, Kati Jarck, Thorsten Koch, Jeff
  Linderoth, et~al.
\newblock Miplib 2017: data-driven compilation of the 6th mixed-integer
  programming library.
\newblock {\em Mathematical Programming Computation}, 13(3):443--490, 2021.

\bibitem[\protect\citeauthoryear{Gomes \bgroup \em et al.\egroup
  }{2008}]{gomes2008connections}
Carla~P Gomes, Willem-Jan Van~Hoeve, and Ashish Sabharwal.
\newblock Connections in networks: A hybrid approach.
\newblock In {\em Integration of AI and OR Techniques in Constraint Programming
  for Combinatorial Optimization Problems: 5th International Conference, CPAIOR
  2008 Paris, France, May 20-23, 2008 Proceedings 5}, pages 303--307. Springer,
  2008.

\bibitem[\protect\citeauthoryear{{Gurobi Optimization, LLC}}{2024}]{gurobi}
{Gurobi Optimization, LLC}.
\newblock {Gurobi Optimizer Reference Manual}.
\newblock \url{https://www.gurobi.com}, 2024.
\newblock Accessed: 2024-01-01.

\bibitem[\protect\citeauthoryear{Guyon \bgroup \em et al.\egroup
  }{2010}]{guyon2010cut}
Olivier Guyon, Pierre Lemaire, {\'E}ric Pinson, and David Rivreau.
\newblock Cut generation for an integrated employee timetabling and production
  scheduling problem.
\newblock {\em European Journal of Operational Research}, 201(2):557--567,
  2010.

\bibitem[\protect\citeauthoryear{He \bgroup \em et al.\egroup
  }{2014}]{he2014learning}
He~He, Hal Daume~III, and Jason~M Eisner.
\newblock Learning to search in branch and bound algorithms.
\newblock {\em Advances in neural information processing systems}, 27, 2014.

\bibitem[\protect\citeauthoryear{Hewitt \bgroup \em et al.\egroup
  }{2010}]{hewitt2010combining}
Mike Hewitt, George~L Nemhauser, and Martin~WP Savelsbergh.
\newblock Combining exact and heuristic approaches for the capacitated
  fixed-charge network flow problem.
\newblock {\em INFORMS Journal on Computing}, 22(2):314--325, 2010.

\bibitem[\protect\citeauthoryear{Huang \bgroup \em et al.\egroup
  }{2022}]{huang2022learning}
Zeren Huang, Kerong Wang, Furui Liu, Hui-Ling Zhen, Weinan Zhang, Mingxuan
  Yuan, Jianye Hao, Yong Yu, and Jun Wang.
\newblock Learning to select cuts for efficient mixed-integer programming.
\newblock {\em Pattern Recognition}, 123:108353, 2022.

\bibitem[\protect\citeauthoryear{J{\"u}nger \bgroup \em et al.\egroup
  }{2009}]{junger200950}
Michael J{\"u}nger, Thomas~M Liebling, Denis Naddef, George~L Nemhauser,
  William~R Pulleyblank, Gerhard Reinelt, Giovanni Rinaldi, and Laurence~A
  Wolsey.
\newblock {\em 50 Years of integer programming 1958-2008: From the early years
  to the state-of-the-art}.
\newblock Springer Science \& Business Media, 2009.

\bibitem[\protect\citeauthoryear{Kingma and Ba}{2014}]{kingma2014adam}
Diederik~P Kingma and Jimmy Ba.
\newblock Adam: A method for stochastic optimization.
\newblock {\em arXiv preprint arXiv:1412.6980}, 2014.

\bibitem[\protect\citeauthoryear{Kipf and Welling}{2016}]{kipf2016semi}
Thomas~N Kipf and Max Welling.
\newblock Semi-supervised classification with graph convolutional networks.
\newblock {\em arXiv preprint arXiv:1609.02907}, 2016.

\bibitem[\protect\citeauthoryear{Labassi \bgroup \em et al.\egroup
  }{2022}]{labassi2022learning}
Abdel~Ghani Labassi, Didier Ch{\'e}telat, and Andrea Lodi.
\newblock Learning to compare nodes in branch and bound with graph neural
  networks.
\newblock {\em arXiv preprint arXiv:2210.16934}, 2022.

\bibitem[\protect\citeauthoryear{Land and Doig}{2010}]{land2010automatic}
Ailsa~H Land and Alison~G Doig.
\newblock {\em An automatic method for solving discrete programming problems}.
\newblock Springer, 2010.

\bibitem[\protect\citeauthoryear{Nair \bgroup \em et al.\egroup
  }{2020}]{nair2020solving}
Vinod Nair, Sergey Bartunov, Felix Gimeno, Ingrid Von~Glehn, Pawel Lichocki,
  Ivan Lobov, Brendan O'Donoghue, Nicolas Sonnerat, Christian Tjandraatmadja,
  Pengming Wang, et~al.
\newblock Solving mixed integer programs using neural networks.
\newblock {\em arXiv preprint arXiv:2012.13349}, 2020.

\bibitem[\protect\citeauthoryear{Paschos}{2014}]{paschos2014applications}
Vangelis~Th Paschos.
\newblock {\em Applications of combinatorial optimization}, volume~3.
\newblock John Wiley \& Sons, 2014.

\bibitem[\protect\citeauthoryear{Paszke \bgroup \em et al.\egroup
  }{2019}]{paszke2019pytorch}
Adam Paszke, Sam Gross, Francisco Massa, Adam Lerer, James Bradbury, Gregory
  Chanan, Trevor Killeen, Zeming Lin, Natalia Gimelshein, Luca Antiga, et~al.
\newblock Pytorch: An imperative style, high-performance deep learning library.
\newblock {\em Advances in neural information processing systems}, 32, 2019.

\bibitem[\protect\citeauthoryear{Paulus \bgroup \em et al.\egroup
  }{2022}]{paulus2022learning}
Max~B Paulus, Giulia Zarpellon, Andreas Krause, Laurent Charlin, and Chris
  Maddison.
\newblock Learning to cut by looking ahead: Cutting plane selection via
  imitation learning.
\newblock In {\em International conference on machine learning}, pages
  17584--17600. PMLR, 2022.

\bibitem[\protect\citeauthoryear{Scarselli \bgroup \em et al.\egroup
  }{2008}]{scarselli2008graph}
Franco Scarselli, Marco Gori, Ah~Chung Tsoi, Markus Hagenbuchner, and Gabriele
  Monfardini.
\newblock The graph neural network model.
\newblock {\em IEEE transactions on neural networks}, 20(1):61--80, 2008.

\bibitem[\protect\citeauthoryear{Schulman \bgroup \em et al.\egroup
  }{2017}]{schulman2017proximal}
John Schulman, Filip Wolski, Prafulla Dhariwal, Alec Radford, and Oleg Klimov.
\newblock Proximal policy optimization algorithms.
\newblock {\em arXiv preprint arXiv:1707.06347}, 2017.

\bibitem[\protect\citeauthoryear{Tang \bgroup \em et al.\egroup
  }{2020}]{tang2020reinforcement}
Yunhao Tang, Shipra Agrawal, and Yuri Faenza.
\newblock Reinforcement learning for integer programming: Learning to cut.
\newblock In {\em International conference on machine learning}, pages
  9367--9376. PMLR, 2020.

\bibitem[\protect\citeauthoryear{Turner \bgroup \em et al.\egroup
  }{2022}]{turner2022adaptive}
Mark Turner, Thorsten Koch, Felipe Serrano, and Michael Winkler.
\newblock Adaptive cut selection in mixed-integer linear programming.
\newblock {\em arXiv preprint arXiv:2202.10962}, 2022.

\bibitem[\protect\citeauthoryear{Turner \bgroup \em et al.\egroup
  }{2023}]{turner2023cutting}
Mark Turner, Timo Berthold, Mathieu Besan{\c{c}}on, and Thorsten Koch.
\newblock Cutting plane selection with analytic centers and multiregression.
\newblock In {\em International Conference on Integration of Constraint
  Programming, Artificial Intelligence, and Operations Research}, pages 52--68.
  Springer, 2023.

\bibitem[\protect\citeauthoryear{Vaswani \bgroup \em et al.\egroup
  }{2017}]{vaswani2017attention}
Ashish Vaswani, Noam Shazeer, Niki Parmar, Jakob Uszkoreit, Llion Jones,
  Aidan~N Gomez, {\L}ukasz Kaiser, and Illia Polosukhin.
\newblock Attention is all you need.
\newblock {\em Advances in neural information processing systems}, 30, 2017.

\bibitem[\protect\citeauthoryear{Wang \bgroup \em et al.\egroup
  }{2023}]{wang2023learning}
Zhihai Wang, Xijun Li, Jie Wang, Yufei Kuang, Mingxuan Yuan, Jia Zeng, Yongdong
  Zhang, and Feng Wu.
\newblock Learning cut selection for mixed-integer linear programming via
  hierarchical sequence model.
\newblock {\em arXiv preprint arXiv:2302.00244}, 2023.

\bibitem[\protect\citeauthoryear{Wesselmann and
  Stuhl}{2012}]{wesselmann2012implementing}
Franz Wesselmann and Uwe Stuhl.
\newblock Implementing cutting plane management and selection techniques.
\newblock In {\em Technical Report}. University of Paderborn, 2012.

\bibitem[\protect\citeauthoryear{Zhang \bgroup \em et al.\egroup
  }{2024}]{zhanglearning}
Sijia Zhang, Fei Shang, Feng Wu, Shaoang Li, and Xiang-Yang Li.
\newblock Learning node selection via tripartite graph representation in mixed
  integer linear programming, 2024.

\end{thebibliography}

\clearpage
\appendix
\section{Related Work}
\textbf{Cut selection in SCIP.} State-of-the-art solvers like SCIP \cite{bestuzheva2021scip} can generate (or separate) various families of cutting planes to populate a cut pool \cite{achterberg2007constraint}. In SCIP, various parameters are tuned to work in concert and control which types of cutting planes are separated and at what frequency, and also which ones are added to the formulation~\cite{balcan2021sample,balcan2022structural}. Specifically, given a pool of available cuts, a parametrized scoring function ranks them, and a cut attaining maximum score is selected to be part of $S^k$. The rest of the cut pool is filtered for parallelism and quality; remaining cuts are sorted again based on their initial score; and the next score-maximizing cut is selected, and so on, continuing until a pre-specified quota of cuts has been chosen for $S^k$, which is finally added to the LP relaxation. At its core, then, cut selection proceeds by choosing one cut at a time and is governed by a scoring function, which in SCIP is a weighted sum of different scores from the MIP literature (integer support, objective parallelism, efficacy, directed cutoff), developed to empirically assess the potential quality of a given cut. 

\textbf{Score-based policy(SBP).} Tang et al.~\shortcite{tang2020reinforcement} represents the first work to combine cut plane selection with machine learning, focusing on learning how to choose cut planes. The paper models the step of selecting cut planes in mixed integer programming problems as a Markov Decision Process (MDP), using a Reinforcement Learning (RL) agent as a submodule within a branch-and-cut algorithm. This work demonstrates the capability of reinforcement learning in improving cut plane selection algorithms and opens a new area of research. Subsequent work by Huang et al.~\shortcite{huang2022learning} proposes a cut ranking method for cut selection in a multiple-instance learning setup. It employs a neural network to score different candidate cuts for the next step, consistently choosing the cut plane with a higher score. Paulus et al.~\shortcite{paulus2022learning} mimics pre-computed cut selection rules, a costly yet valuable strong branching-like method, aiming to try each choice and obtain the cut set that maximizes the improvement of bounds. It encodes the system into a tripartite graph, where nodes contain vector features representing variables, constraints, and available cuts, offering a new direction for cut plane selection. 


\textbf{Hierarchical sequence method (HEM).} Wang et al.~\shortcite{wang2023learning} emphasize the significance of the number of cut planes, the collective impact of different cuts, and the sequence in which they are added in the context of cut plane selection. To address these factors, they propose a two-tier network for reinforcement learning. The upper network is responsible for outputting the number of cut planes to select, while the lower network employs a pointer mechanism to determine both the specific cut planes to be added and the order in which they should be incorporated. This hierarchical approach effectively captures the complex interactions among cuts, enhancing the overall efficiency of the selection process. By integrating these elements, HEM provides a sophisticated framework for optimizing cut plane selection in mixed-integer programming.

\section{List of Symbols}\label{appendix:symbols}
We provide a list of the key symbols used throughout the paper. These symbols represent various components of the mixed-integer programming (MIP) problem, the reinforcement learning (RL) framework, and the network architecture. The symbols are categorized into different sections based on their function in the model. For easy reference, Table~\ref{tab:symbols} provides the definitions of these symbols.

\begin{table*}[htbp]
\centering
\caption{List of Symbols Used in the Paper}
\label{tab:symbols}
\renewcommand{\arraystretch}{1.5}
\begin{tabular}{c|l}
\toprule
\textbf{Symbol} & \textbf{Description} \\
\midrule
\(\mathcal{S}\) & State space, representing the search tree or bipartite graph \\
\(\mathcal{A}\) & Action space, representing all possible selections and orderings of cutting planes \\
\(\mathcal{T}\) & Transition function, mapping from state and action to the next state \\
\(\mathcal{R}\) & Reward function, evaluating the improvement over the baseline \\
\(\mathbf{f}_i\) & Feature vector for the \(i\)-th node in the bipartite graph \\
\(\mathbf{h}_i\) & Embedded feature vector for node \(i\) in the graph \\
\(\mathbf{h}_i^{(\text{final})}\) & Final embedded feature vector for node \(i\) after graph convolution \\
\(\mathbf{h}_P\) & Pooled feature vector for the candidate cutting planes \(P\) \\
\(\mathbf{H}_{\text{transformer}}\) & Hidden states output by the transformer sequence model \\
\(p_i\) & Probability of selecting the \(i\)-th cutting plane, output by the policy network \\
\(y_i\) & Binary variable indicating the selection status of cutting plane \(c_i\) \\
\(\pi\) & Permutation of selected cutting planes \\
\(\text{gap}_t\) & Gap at iteration \(t\), defined as the relative difference between Best Primal and Best Feasible \\
\(\Delta \text{gap}_t\) & Change in gap from iteration \(t-1\) to \(t\) \\
\(\text{Best Primal}\) & The best solution found by the LP relaxation \\
\(\text{Best Feasible}\) & The best feasible solution found by the solver \\
\(T_{\text{baseline}}\) & Solving time for the baseline (e.g., SCIP) \\
\(T_{\text{method}}\) & Solving time for the proposed method \\
\(\sigma\) & Sigmoid activation function used in the policy network \\
\(W_{\text{policy}}\) & Weight matrix used in the policy network \\
\(b_{\text{policy}}\) & Bias term in the policy network \\
\midrule
\end{tabular}
\end{table*}

\section{Theoretical Analysis}
\subsection{Explanation of Action Space Complexity} \label{appendix:actionspace}

For a set with \( n \) elements, each element can either be selected or not selected, and the order of selection matters. The total number of possible combinations is calculated as follows:

1. Selection: Each element has two choices: to be selected or not selected. Thus, for \( n \) elements, there are \( 2^n \) ways to choose elements.
2. Permutation: For each selection of \(k\) elements, there are \(k!\) ways to arrange them. Combining these two factors, the total number of permutations for all possible selections is:
\[
\text{Total} = \sum_{k=0}^{n} \binom{n}{k} \cdot k! = \sum_{k=0}^{n} \frac{n!}{(n-k)!}
\]
To analyze the complexity of this calculation, we note that the sum has \( n+1 \) terms, and each term involves the computation of \( n! \). Although the complexity of individual terms varies depending on \( k \), the overall complexity is dominated by the largest term, which is \( n! \). Thus, the overall complexity can be approximated as \( O(n \cdot n!) \).

\section{More Details of GCS}
\subsection{Node and Edge Features in GCS Bipartite Graph Representation}\label{appendix:graph_feature}
The GCS approach uses an advanced bipartite graph to capture the branch-and-cut process, with features for nodes and edges representing the relationships between variables, constraints, and cuts.
A list of the features included in our bipartite state representation is given as Table~\ref{table:feature}.
This structured representation helps capture the dynamics of the branch-and-cut process, enabling more informed decision-making for cut selection.

\begin{table*}[htbp]
    \centering
    \caption{Description of node and edge features in the GCS bipartite graph representation}
    \label{table:feature}
    \renewcommand{\arraystretch}{1.5} 
    \begin{tabular}{>{\centering\arraybackslash}m{3cm}>{\centering\arraybackslash}m{4cm}>{\centering\arraybackslash}m{7cm}}
        \toprule
        \textbf{Category} & \textbf{Feature} & \textbf{Description} \\
        \midrule
        \multirow{4}{3cm}{\centering variable vertex} 
            & lb & Lower bound. \\ \cline{2-3}
            & ub & Upper bound. \\ \cline{2-3}
            & objective\_coef & Coefficient in the objective function. \\ \cline{2-3}
            & var\_type &  Type (binary, integer, continuous).\\
        \midrule
        \multirow{4}{3cm}{\centering constraint vertex} 
            & rhs & Right-hand side of the constraint. \\ \cline{2-3}
            & lhs & Left-hand side of the constraint. \\ \cline{2-3}
            & cons\_type & Constraint type feature (eq, geq).\\ \cline{2-3}
            & non-zero\_num  & Number of non-zero coefficients in the constraint. \\ 
        \midrule
        \multirow{3}{3cm}{\centering cutting plane vertex} 
            & addition time & Time when the cut was added. \\ \cline{2-3}
            & improvement effect & Effect of the cut on improving the bound. \\ \cline{2-3}
            & parallelism & Parallelism with the objective function. \\ 
        \midrule
        \multirow{8}{3cm}{\centering candidate cutting plane vertex} 
             & parallelism & Parallelism with the objective function. \\  \cline{2-3}
            & efficacy &  Euclidean distance of the cut hyperplane to the current LP solution. \\ \cline{2-3}
            & support &  Proportion of non-zero coefficients of the cut. \\ \cline{2-3}
            & integral support &  Proportion of non-zero coefficients with respect to integer variables of the cut. \\ \cline{2-3}
            & normalized violation &  Violation of the cut to the current LP solution $\max\{0,\frac{\alpha^T x_{LP}^*-\beta}{|\beta|}\}$. \\ 
        \midrule
        \multirow{1}{3cm}{\centering con-var edge} 
            & coef & Constraint coefficient. \\ 
        \midrule
        \multirow{2}{3cm}{\centering cut-var edge} 
            & coef & Constraint coefficient. \\ \cline{2-3}
            & branch\_cons & When cuts are added, the branching constraint (sign, rhs) for the node is included. \\ 
        \bottomrule
    \end{tabular}
\end{table*}

\subsection{Reinforcement Learning Training Process for GCS}
\label{appendix:train}
To train the reinforcement learning (RL) policy for the GCS method, we utilize the Proximal Policy Optimization (PPO) algorithm. This training framework integrates the MIP instances into a policy gradient setting, where the objective is to optimize the policy $\pi_{\theta}$ for selecting cuts to improve solver performance. The training process involves: (1) Computing immediate rewards based on the gaps between the current solution and the optimal one; (2) Scaling and discounting the rewards to account for future returns; (3) Updating the policy and value networks by minimizing the respective losses using gradient descent.

The following pseudocode outlines the steps in the training process.

\begin{algorithm}[ht]
    \caption{Training Process using PPO for Cut Selection}
    \label{alg:train_ppo}
    \textbf{Initialize}: Policy Network $\pi_{\theta}$, Value Network $V_{\phi}$, Optimizers $\mathcal{O}_{\pi}$ and $\mathcal{O}_{V}$, states $S$, actions $A$, rewards $R$, and the training parameters such as $\gamma$ (discount factor) and $\epsilon$ (PPO clip threshold). \\
    \begin{algorithmic}[1] 
        \STATE \textbf{For each episode:}
        \STATE \quad Initialize episode data structures (states, actions, immediate rewards)
        \FOR{each time step in episode}
            \STATE Sample state $s_t$ from environment
            \STATE Select action $a_t$ using the policy network $\pi_{\theta}(s_t)$
            \STATE Observe reward $r_t$ based on the action
            \STATE Store $(s_t, a_t, r_t)$ in episode data
        \ENDFOR
        \STATE Compute discounted rewards $R_t$ using $\gamma$ and final reward
        \STATE Calculate advantage estimates using the value network $V_{\phi}(s_t)$
        
        \STATE \textbf{Policy Update:}
        \FOR{several epochs}
            \FOR{each sample $(s_t, a_t)$ in the batch}
                \STATE Compute log-probabilities $\log \pi_{\theta}(a_t|s_t)$
                \STATE Compute the ratio $r_t = \frac{\pi_{\theta}(a_t|s_t)}{\pi_{\theta_{\text{old}}}(a_t|s_t)}$
                \STATE Compute surrogate losses: $L_{\text{PPO}} = \min(r_t A_t, \text{clip}(r_t, 1-\epsilon, 1+\epsilon) A_t)$
                \STATE Compute policy loss: $L_{\pi} = -\mathbb{E}[L_{\text{PPO}}]$
            \ENDFOR
            \STATE Update policy network: $\theta \leftarrow \theta - \alpha_{\pi} \nabla_{\theta} L_{\pi}$
        \ENDFOR
        
        \STATE \textbf{Value Update:}
        \STATE Compute value loss $L_V = \mathbb{E}[(V_{\phi}(s_t) - R_t)^2]$
        \STATE Update value network: $\phi \leftarrow \phi - \alpha_V \nabla_{\phi} L_V$
    \end{algorithmic}
\end{algorithm}

The PPO algorithm is used to update the policy and value networks. The policy network selects cutting planes based on the state of the MIP solver, while the value network estimates the expected return from a given state. These networks are optimized using standard backpropagation and gradient descent. The PPO approach ensures that the updates do not change the policy too drastically, which helps stabilize the learning process. The training process is repeated over multiple episodes, where each episode corresponds to a different MIP instance from the problem set. After training, the policy network is evaluated for its performance in terms of solving time and node count compared to standard solver configurations.



\section{More Results}
\subsection{Dataset Details} \label{appendix:dataset_details}
\begin{table}[ht]
\centering
\caption{Number of instances used for training and testing across datasets.}
\begin{tabular}{lcc}
\toprule
\textbf{Dataset} & \textbf{Training Instances} & \textbf{Testing Instances} \\ \midrule
SetCover         & 800                         & 200                         \\ 
MIS              & 800                         & 200                         \\ 
GISP             & 200                         & 50                          \\ 
FCMCNF           & 200                         & 50                          \\ 
MIK              & 72                          & 18                          \\ 
CORLAT           & 200                         & 50                          \\ \bottomrule
\end{tabular}
\label{tab:dataset_split}
\end{table}
For all datasets used in this study, we followed a consistent strategy of splitting the available instances into training and testing sets, with 80\% of the instances allocated for training and 20\% for testing. This approach ensures a balanced evaluation while allowing the model to generalize well to unseen data. A summary of the datasets, including the number of instances used for both training and testing, is provided in Table~\ref{tab:dataset_split}.

Each dataset represents a different type of optimization problem and was carefully selected to provide a diverse set of challenges. For example, the \textit{SetCover} dataset consists of instances from the set cover problem, while the \textit{MIS} dataset contains instances from the maximum independent set problem. The \textit{GISP} and \textit{FCMCNF} datasets represent graph-based optimization problems, and the \textit{MIK} and \textit{CORLAT} datasets are drawn from industrial and combinatorial optimization domains. The wide range of problems covered ensures that the models tested in this study can handle various real-world optimization tasks.
\subsection{Dataset Statistics} \label{appendix:dataset_stat}

The statistics of the datasets used in this study are summarized in Table~\ref{tab:dataset_stats}. Each dataset is characterized by the average number of constraints ($m$), variables ($n$), graph size (nodes and edges), and the average number of candidate cuts. Additionally, the table provides the interface time required for GCS decision-making on each dataset. These metrics illustrate the varying complexities of the datasets, with larger instances like GISP and FCMCNF requiring more time for processing compared to smaller instances like MIK. Notably, the number of candidate cuts and interface time highlight the computational challenges posed by different datasets.


\begin{table*}[t]
\centering
\caption{Statistical description of datasets and interface time for GCS decision-making.}
\label{tab:dataset_stats}
\renewcommand{\arraystretch}{1.3} 
\begin{tabular}{lcccccccc}
\toprule
\textbf{Feature} & \textbf{SetCover} & \textbf{MIS} & \textbf{GISP(p=95)} & \textbf{FCMCNF(p=21)} & \textbf{MIK} & \textbf{CORLAT} \\ 
\midrule
$m$ (Constraints)  & 1000 & 500  & 2371     & 585    & 413     & 466      \\ 
$n$ (Variables)  & 500   & 1953   & 1305& 3168    & 346      & 486    \\ 
Nodes       & 2886     & 3089        & 10896     & 5753    & 3412    & 4993            \\ 
Edges & 31359       & 8844     & 20854  & 17909     & 27859   & 32584           \\ 
Avg. Candidate Cuts         & 700 & 55 & 43 & 109 & 51 & 65 \\ 
\bottomrule
\end{tabular}
\end{table*}
\begin{table*}[t]
    \centering
    \scriptsize
    \caption{Solve time (s) on selected MIPLIB 2017 instances.}
    \begin{tabular}{lcccccccc}
        \toprule
        \textbf{Method} & \textbf{neos-18} & \textbf{piperout-d27} & \textbf{neos-555424} & \textbf{30\_70\_45\_05\_100} & \textbf{30\_70\_45\_095\_100} & \textbf{acc-tight2} & \textbf{neos-933562} & \textbf{istanbul} \\
        \midrule
        SCIP & 14.7 & 85.7 & 600 & 600 & 202.1 & 24.0 & 600 & \textbf{122.9} \\ 
        SBP & 15.5 & 83.9 & 600 & 600 & 192.2 & 21.9 & 600 & 137.4 \\ 
        HEM & 14.0 & 83.3 & 600 & 600 & 195.9 & 20.5 & 600 & 140.6 \\ 
        \midrule
        GCS (Ours) & \textbf{13.3} & \textbf{67.1} & 600 & \textbf{518.7} & \textbf{176.4} & \textbf{19.5} & 600 & 124.7 \\ 
        \bottomrule
    \end{tabular}
    \label{tab:miplib_results}
\end{table*}

\subsection{Additional Results on MIPLIB 2017} \label{appendix:miplib}
MIPLIB 2017 benchmark dataset~\cite{gleixner2021miplib} contains MIP instances from various application domains and has long been a standard benchmark for MIP solvers~\cite{nair2020solving,turner2022adaptive,gleixner2021miplib}. Previous studies ~\cite{turner2022adaptive} have shown that directly learning over the entire MIPLIB can be extremely challenging due to the heterogeneity of the instances, and machine learning methods face difficulties in selecting similar instances. 

Following the method of Wang et al.~\shortcite{wang2023learning}, we selected the MIPLIB support case dataset, containing 40 instances split into 80\% for training and 20\% for testing. Notably, the similarity between instances was determined using 100 human-designed instance features~\cite{gleixner2021miplib}. The instances in the dataset include supportcase40, 30\_70\_45\_05\_100, 30\_70\_45\_095\_100, acc-tight2, acc-tight4, acc-tight5, comp07-2idx, comp08-2idx, comp12-2idx, comp21-2idx, decomp1, decomp2, gus-sch, istanbul-no-cutoff, mkc, mkc1, neos-555343, neos-555424, neos-738098, neos-872648, neos-933562, neos-933638, neos-933966, neos-935234, neos-935769, neos-983171, neos-1330346, neos-1337307, neos-1396125, neos-3209462-rhin, neos-3755335-nizao, neos-3759587-noosa, neos-4300652-rahue, neos18, physiciansched6-1, physiciansched6-2, piperout-d27, qiu, reblock354, and supportcase37. A maximum solve time of 600 seconds was set to account for varying problem difficulty. Results for eight test instances are summarized in Table~\ref{tab:miplib_results}.

These results demonstrate that the GCS method consistently outperforms baseline approaches such as SCIP, SBP, and HEM on challenging MIP instances. The improvements are particularly noticeable in instances requiring complex cut selection, showcasing the effectiveness of leveraging global graph-based information for decision-making.

\end{document}